\documentclass{llncs}

\usepackage{graphicx}
\usepackage{amsmath}
\usepackage{amssymb}

\renewcommand{\emptyset}{\varnothing}

\renewcommand{\geq}{\geqslant}

\def\CPP{C\raise1.5pt\hbox{\footnotesize ++}}
\def\CPPsmall{C\raise1.3pt\hbox{\tiny ++}}

\def\DOM{\mathcal{D}}

\def\Seq#1{#1^\circ}
\def\Par#1{#1^\cap}
\def\Gfp#1{#1^\omega}

\def\Gen#1{\mathtt{Gen}(#1)}

\def\GI{\texttt{GIco}}

\newcommand{\LIMF}[1]{#1\mathop{\uparrow\omega}}

\title{Enhancing Constraint Propagation with Composition Operators~\thanks{This work has been supported 
    by the IST project COCONUT from the European Community. An extended version of this paper will be
  submitted to the CP'2001 conference.
    E-mails: $\{$granvilliers,monfroy$\}$@irin.univ-nantes.fr}}
\author{Laurent Granvilliers
\and Eric Monfroy} 
\institute{Computer Science Research Institute {\scriptsize (IRIN)}\\
 University of Nantes\\
44322 Nantes cedex 3, France
}

\begin{document}
\maketitle

\begin{abstract}
  Constraint propagation is a general algorithmic approach for
  pruning the search space of a CSP. In a uniform way,
  K.~R.~Apt~\cite{AptTCS1999} has defined a computation as an iteration of
  reduction functions over a domain. In~\cite{AptTOPLAS2001} he has
  also demonstrated the need for integrating static properties of
  reduction functions (commutativity and semi-commutativity) to design
  specialized algorithms such as \texttt{AC3} and \texttt{DAC}. We
  introduce here a set of operators for modeling compositions of
  reduction functions. Two of the major goals are to tackle parallel
  computations, and dynamic behaviours (such as slow
  convergence).
\end{abstract}

\section{Introduction}\label{sec:intro}
A Constraint Satisfaction Problem (CSP) is defined by a set of
domains, a set of variables, and a set of constraints. Solving a CSP
consists in finding assignments of variables (i.e., values from their
domains) that satisfy constraints. Since this problem is NP-hard,
preprocessing techniques have been implemented to prune domains
(search space) before backtracking, e.g., filtering algorithms based
upon consistency properties of subsets of
constraints~\cite{MackworthAI1977}.  Constraint propagation is a
generic term for these techniques.

Recently, K.~R.~Apt~\cite{AptTCS1999} has proposed a unified framework for
constraint propagation. The solving process is defined as a chaotic
iteration~\cite{ChazanLAA1969}, i.e., an iteration of reduction
functions over domains. Under well-chosen properties of domains (e.g.,
partial ordering, well-foundedness) and functions (e.g., monotonicity,
contractance), iteration-based algorithms are shown to be finite and
confluent. Further refinements have been devised
in~\cite{AptTOPLAS2001} to tackle strategies based
on additional properties of functions (such as \texttt{AC3} with
idempotence and commutativity). Hence, specializations of component functions of the
generic iteration algorithm have been implemented to tune the order of
applications of functions.

In this paper we introduce new properties of functions (i.e.,
strongness, independency, and redundancy) to tackle parallel
computations~\cite{GranvilliersIPL2000}, and dynamic behaviours such
as slow convergence~\cite{LhommeJLP1998}. Furthermore, the dynamic
nature of such strategies led us to define the notion of composition
operator.  Basically, a composition operator is a local strategy for
applying functions: it implements combinators for sequential or
parallel computation, and for computation of closures.  An iteration
(preserving the semantics ---termination, confluence---) is then
defined as a sequence of application of composition operators.  This
approach provides flexibility (definition of components for creating
composition operators) and generality (single iteration algorithm).

Using composition operators, we then present several strategies for
modeling heuristics and properties of well-known constraint solvers:
priorities of constraints in the finite domain solver
\texttt{Choco}~\cite{Choco}, slow convergence arising in interval
narrowing~\cite{LhommeJLP1998}, efficient heuristics for interval
narrowing~\cite{GranvilliersRC2001} as is done in
\texttt{Numerica}~\cite{Numerica}, and finally, parallel
computations~\cite{GranvilliersIPL2000}. Essentially, each strategy
dynamically creates and applies composition operators until a
fixed-point is computed.

The outline of this paper is the following.  A basic  constraint
propagation framework is introduced in Section~\ref{sec:chaotic}. Our
new framework based on composition operators is described in
Section~\ref{sec:extension}. In Section~\ref{sec:spec}, some existing
strategies are shown to fit in our model.


\section{Constraint Propagation}\label{sec:chaotic}
Our aim is not to be as general as in~\cite{AptTOPLAS2001}. We
consider here contracting and monotonic functions on a finite
semilattice ordered by set inclusion. Such a domain is used to address the
decoupling of reductions arising in parallel
computations (see Section~\ref{sec:extension}). Note that the results
given here hold under these assumptions.

The \emph{computation domain} is a finite semilattice
$(\DOM,\subseteq,\cap)$, i.e., a partially ordered set
$(\DOM,\subseteq)$ in which every nonempty finite subset has a
greatest lower bound (an $\inf$ element). The ordering
$\subseteq$ corresponds to set inclusion. The meet operation
is the set intersection $\cap$, and the bottom element is the empty
set.  Note that in most constraint solvers, the computation
domain is a finite lattice, i.e., finite domains such as Booleans,
Integers, floating-point intervals.

\begin{definition}[Reduction function]
Consider a function $f$ on $\DOM$.
\begin{itemize}
\item $f$ is contracting if $\forall x\in\DOM,\ f(x)\subseteq x$;
\item $f$ is monotonic if $\forall x,y\in\DOM,\ x\subseteq y\Rightarrow f(x)\subseteq f(y)$.
\end{itemize}
A \emph{reduction function on $\DOM$} is a contracting and monotonic
function on $\DOM$.
\end{definition}

In the following we consider a finite set of reduction functions
$F=\{f_1,\dots,f_k\}$ on $\DOM$. 
\begin{definition}[Iteration]\label{def:i}
  Given an element $d\in\DOM$, an iteration of
  $F$ over $d$ is an infinite sequence of values $d_0,d_1,\dots$ defined
  inductively by:
$$
\begin{array}{lcl}
  d_0 & \hbox{\emph{\texttt{:=}}} & d\\[1mm]
  d_i & \hbox{\emph{\texttt{:=}}} & f_{j_i}(d_{i-1})\ \ i\geq 1
\end{array}
$$
where $j_i$ is an element of $[1,\dots,k]$.
A sequence $d_0\supseteq d_1\supseteq\cdots$ of elements from $\DOM$
\emph{stabilizes} at $e$ if for some $j\geq 0$ we have $d_i=e$ for $i\geq j$.
\end{definition}

We have the following stabilization lemma.
\begin{lemma}\label{lem:stab}
Suppose that an iteration of $F$ over $d$ stabilizes at a common
fixed-point $e$ of the functions from $F$. Then, $e$ is the greatest
common fixed-point of the functions from $F$ that is included in $d$,
i.e., $(\cap_{i=1}^{k}f_i)\uparrow\omega(d)$.
\begin{proof}
See~\cite{AptTOPLAS2001}. It follows from the monotonicity of the
reduction functions.
\end{proof}
\end{lemma}

The generic iteration algorithm (\texttt{GI}) defined by K.~R.~Apt
in~\cite{AptTOPLAS2001} is given in Table~\ref{tab:gi}. The
correctness of \texttt{GI} is stated by Theorem~\ref{theo:gi}.

\begin{table}
\caption{Generic Iteration Algorithm based on Reduction Functions.}
\label{tab:gi}
\begin{tabbing}
\hspace*{12mm}\=\hspace*{5mm}\=\hspace*{5mm}\=\kill
\> \texttt{GI} ($F$: set of reduction functions on $\DOM$ ; $d$: element of $\DOM$): $\DOM$\\
\> \texttt{begin}\\
\> \> $G$ \texttt{:=} $F$\\
\> \> \texttt{while} $G\neq\emptyset$ \texttt{do}\\
\> \> \> \texttt{choose} $g\in G$\\
\> \> \> $G$ \texttt{:=} $G - \{g\}$\\
\> \> \> $G$ \texttt{:=} $G\cup \texttt{update}\,(G,g,d)$\\
\> \> \> $d$ \texttt{:=} $g(d)$\\
\> \> \texttt{od}\\
\> \> \texttt{return} $d$\\
\> \texttt{end}\\[2mm]
\> where for all $G,g,d$ the set of functions $\texttt{update}\,(G,g,d)$
is such that\\
  \> A. \> $\{f\in F-G\mid f(d)=d\ \wedge\ fg(d)\neq g(d)\}\subseteq\texttt{update}\,(G,g,d)$\\
  \> B. \> $g(d)=d$ implies that $\texttt{update}\,(G,g,d)=\emptyset$\\
  \> C. \> $gg(d)\neq g(d)$ implies that $g\in\texttt{update}\,(G,g,d)$
\end{tabbing}
\end{table}

\begin{theorem}\label{theo:gi}
Every execution of \emph{\texttt{GI}} terminates and computes in $d$
the greatest common fixed-point of the functions from $F$.
\begin{proof}
See~\cite{AptTOPLAS2001}. It follows from the monotonicity and
contractance of the reduction functions, and the well-foundedness of
the ordering of the semilattice.
\end{proof}
\end{theorem}

\section{Composition Operators}\label{sec:extension}
We restrict our attention to simple domains. The generalization of our
results to compound domains, i.e., considering $k$-ary functions on
$(\DOM_1,\dots,\DOM_k)$, is straightforward.

\subsection{Definitions}
Consider a finite set of functions $F=\{f_1,\dots,f_k\}$ on $\DOM$. We introduce a set of composition operations on $F$ as follows:
$$
\begin{array}{rllll}
\hbox{Sequence:} &~~& \Seq{F} & \hbox{~~denotes the function~~} &
  x\mapsto f_1f_2\dots f_k(x)\\
\hbox{Closure:} && \Gfp{F} & &
  x\mapsto (\cap_{i=1}^{k}f_i)\uparrow\omega(x)\\
\hbox{Decoupling:} && \Par{F} & &
  x\mapsto f_1(x)\cap\cdots\cap f_k(x)
\end{array}
$$
Note that $F$ is supposed to be ordered since the sequence
operation is not commutative.  This assumption is no longer necessary
for the closure and decoupling operations since the computation of a
fixed-point is a declarative property, and the intersection operation
is commutative.

There are several motivations for introducing such operations in a
generic propagation framework:
\begin{itemize}
\item The sequence operation fixes the order of application of
  the reduction functions. It can be used for computing directional arc
  consistency based on the semi-commutativity property, for modeling
  priorities of solvers, and for implementing heuristics or knowledge
  of solvers about their relative efficiencies.

\item The closure operation allows us to make a closure from a non
  idempotent function, and to describe multi-level algorithms that
  compute fixed-points of different solvers.
  
\item The decoupling operation is essentially used to model
  parallel computations, enforcing different functions
  over the same domain, and then computing the intersection using a
  fold reduction step.
\end{itemize}

The notion of composition operator models a function (a complex
solver) built from composition operations.
\begin{definition}[Composition operator]\label{def:compo}
  Let $F=\{f_1,\dots,f_k\}$ be a finite set of reduction functions.
  A \emph{composition operator on $F$} is a
  function $\DOM\to\DOM$ defined by induction as follows:
$$
\begin{array}{lrll}
& \hbox{\emph{Atomic:}} &~~& f_i \hbox{ is a composition operator for } i=1,\dots,k.\\[3mm]
\multicolumn{4}{l}{\hbox{Given a finite set } \Phi \hbox{ of composition operators on } F,}\\[1mm]
& \hbox{\emph{Sequence:}}   &~~& \Seq{\Phi} \hbox{ is a composition operator;}\\
& \hbox{\emph{Closure:}}    &~~& \Gfp{\Phi} \hbox{ is a composition operator;}\\
& \hbox{\emph{Decoupling:}} &~~& \Par{\Phi} \hbox{ is a composition operator.}
\end{array}
$$ The \emph{generator $\Gen{\phi}$} of a composition operator $\phi$
on $F$ is the subset of functions from $F$ that are ``involved'' in
$\phi$. It is defined inductively by:
\begin{itemize}
\item $\Gen{\phi}=\{\phi\}$ if $\phi$ is an atomic operator from $F$;
\item $\Gen{\phi}=\Gen{\phi_1}\cup\cdots\cup\Gen{\phi_k}$ if
   $\phi=\{\phi_1,\dots,\phi_k\}^\star$ for $\star\in\{\circ,\omega,\cap\}$.
\end{itemize}
\end{definition}

Lemma~\ref{lem:compo} states that a composition operator is also a
reduction function.
\begin{lemma}\label{lem:compo}
Consider a finite set of reduction functions $F=\{f_1,\dots,f_k\}$.
Then, every composition operator on $F$ is
\begin{itemize}
\item[(i)] contracting;
\item[(ii)] monotonic.
\end{itemize}
\begin{proof}
  The proof is done by induction. Every atomic operator
  is contracting and monotonic by definition of a reduction function.
  Now consider a set $\Phi=\{\phi_1,\dots,\phi_k\}$ of contracting
  and monotonic composition operators.
  \begin{itemize}
  \item[(i)] By hypothesis the composition operators from $\Phi$
    are contracting. Then it is immediate to prove the contractance of
    $\Seq{\Phi}$, $\Gfp{\Phi}$, and $\Par{\Phi}$.

  \item[(ii)] Given $x,y\in\DOM$ suppose that $x\subseteq y$.
    \begin{itemize}
    \item[--] Since every $\phi_i$ is supposed to be monotonic, then,
    we have $\phi_k(x)\subseteq\phi_k(y)$, and then,
    $\phi_{k-1}\phi_k(x)\subseteq\phi_{k-1}\phi_k(y)$,
    and so on. It follows that $\Seq{\Phi}$ is monotonic.
       
    \item[--] To prove the monotonicity of $\Gfp{\Phi}$ we
     consider the function $\varphi:\cap_{i=1}^k \phi_i$.
     Then, we prove that $\varphi$ is monotonic (third item).
     It follows that
     $\varphi\uparrow\omega(x)\subseteq\varphi\uparrow\omega(y)$.
     Then it is immediate to prove (by a double inclusion) that the
     set of fixed-points of $\varphi$ coincides with the set of common
     fixed-points of the functions from $\Phi$.
     As a consequence, we have
     $\varphi\uparrow\omega\equiv\Gfp{\Phi}$, that completes the
     proof.
      
    \item[--] For $i=1,\dots,k$, we have $\phi_i(x)\subseteq \phi_i(y)$ 
    by monotonicity of $\phi_i$. It follows that
    $\cap_{i=1}^k \phi_i(x)\subseteq\cap_{i=1}^k \phi_i(y)$, that ends
    the proof.
    \end{itemize}
  \end{itemize}
\end{proof}
\end{lemma}

Lemma~\ref{lem:fp} states that a fixed-point of a composition operator
is a common fixed-point of the functions from its generator. The key idea
is that the application of a composition operator implies that
each reduction function in its generator is applied at least once.
\begin{lemma}\label{lem:fp}
  Consider a finite set of reduction functions $F$ and a composition
  operator $\phi$ on $F$. Then, $e$ is a fixed-point of $\phi$ if and only
  if $e$ is a common fixed-point of the functions from $\Gen{\phi}$.
\begin{proof} We prove by induction 
the equivalence $\phi(e)=e\Leftrightarrow \forall f\in\Gen{\phi}\ f(e)=e$.
It obviously holds for an atomic operator $\phi$. Now consider a set
of composition operators $\Phi=\{\phi_1,\dots,\phi_k\}$ on $F$ and
assume that the equivalence holds for each $\phi_1,\dots,\phi_k$.
If we have $\phi\equiv\Phi^\star$ for
$\star\in\{\circ,\omega,\cap\}$, then it follows:
$$
\begin{array}{lcll}
  \phi(e)=e & \Leftrightarrow & \forall i\in\{1,\dots,k\}\ \phi_i(e)=e
  & \hbox{(immediate result)}\\
            & \Leftrightarrow & \forall i\in\{1,\dots,k\}\ \forall f\in\Gen{\phi_i}\ f(e)=e & \hbox{(induction hypothesis)}\\
\end{array}
$$
Considering $\Gen{\phi}=\cup_i\Gen{\phi_i}$ completes the proof.
\end{proof}
\end{lemma}

The notion of iteration is slightly extended to deal with composition
operators instead of reduction functions.
\begin{definition}[Iteration]\label{def:iter}
  Consider a finite set of reduction functions $F$, and a finite
  set $\Phi=\{\phi_1,\dots,\phi_k\}$ of composition
  operators on $F$. Given an element $d\in\DOM$, an iteration of
  $\Phi$ over $d$ is an infinite sequence of values $d_0,d_1,\dots$ defined
  inductively by:
$$
\begin{array}{lcl}
  d_0 & \hbox{\emph{\texttt{:=}}} & d\\[1mm]
  d_i & \hbox{\emph{\texttt{:=}}} & \phi_{j_i}(d_{i-1})\ \ i\geq 1
\end{array}
$$
where $j_i$ is an element of $[1,\dots,k]$.
\end{definition}

Lemma~\ref{lem:stab} (stabilization lemma from K.~R.~Apt) remains valid,
that follows from the monotonicity of the composition operators
(Lemma~\ref{lem:compo}). Moreover, we have the following, essential
result.
\begin{lemma}\label{lem:fixed}
If an iteration on $\Phi=\{\phi_1,\dots,\phi_k\}$ over $d$ stabilizes at
a common fixed-point $e$ of the functions from $\Phi$,
and $F=\cup_i\Gen{\phi_i}$, then $e=\Gfp{F}(d)$.
\begin{proof} We first prove that $e$ is a common fixed-point of the functions
    from $F$. By hypothesis $e$ is a fixed-point of each $\phi_i\in \Phi$.
    By Lemma~\ref{lem:fp} it follows that $f(e)=e$ for each $f\in
    \Gen{\phi_i}$. The proof is completed since by hypothesis the set of
    generators covers $F$.
  
    We prove now that $e$ is the greatest common fixed-point of the
  functions from $F$. Consider a common fixed-point $e'$ of the
  functions from $F$. It suffices to prove that $e'$ is included in
  every element from the iteration, namely $d_0,d_1,\dots$. It
  obviously holds for $i=0$. Suppose now it holds for $i$, i.e.,
  $e'\subseteq d_i$, and assume that $d_{i+1}=\phi_j(d_i)$ for some
  $j\in[1,\dots,k]$.  By monotonicity of $\phi_j$ we have
  $\phi_j(e')\subseteq \phi_j(d_i)$. By Lemma~\ref{lem:fp} we have
  $\phi_j(e')=e'$, that completes the proof.
\end{proof}
\end{lemma}

\begin{table}[htbp]
\caption{Generic Iteration Algorithm based on Composition Operators.}
\label{tab:gico}
\begin{tabbing}
\hspace*{12mm}\=\hspace*{5mm}\=\hspace*{5mm}\=\kill
\> \GI ($F$: set of reduction functions on $\DOM$ ; $d$: element of $\DOM$): $\DOM$\\
\> \texttt{begin}\\
\> \> $G$ \texttt{:=} $F$\\
\> \> \texttt{while} $G\neq\emptyset$ \texttt{do}\\
\> \> \> $\phi$ \texttt{:=} \texttt{create a composition operator on} $G$\\
\> \> \> $G$ \texttt{:=} $G - \Gen{\phi}$\\
\> \> \> $G$ \texttt{:=} $G\cup \texttt{update}\,(G,\phi,d)$\\
\> \> \> $d$ \texttt{:=} $\phi(d)$\\
\> \> \texttt{od}\\
\> \> \texttt{return} $d$\\
\> \texttt{end}\\[2mm]
\> where for all $G,\phi,d$ the set of functions $\texttt{update}\,(G,\phi,d)$
is such that\\
  \> A. \> $\texttt{upA}\,\texttt{:=}\ \{f\in F-G\mid f(d)=d\ \wedge\ f\phi(d)\neq\phi(d)\}\subseteq\texttt{update}\,(G,\phi,d)$\\
  \> B. \> $\texttt{upB}\,\texttt{:=}\ \phi(d)=d$ implies that $\texttt{update}\,(G,\phi,d)=\emptyset$\\
  \> C. \> $\texttt{upC}\,\texttt{:=}\ \{f\in\Gen{\phi}\mid f\phi(d)\neq\phi(d)\}\subseteq\texttt{update}\,(G,\phi,d)$
\end{tabbing}
\end{table}

We describe now a generic iteration algorithm \GI{} based on
composition operators on a finite set of reduction functions $F$ (see
Table~\ref{tab:gico}). Note that the set of composition operators is
not fixed, since each operator is dynamically created from the set $G$
of active reduction functions, and it is applied only once.
Nevertheless, Theorem~\ref{theo:gico} proves the correctness of \GI{}
with respect to $F$.

\begin{theorem}\label{theo:gico}
Every execution of \emph{\GI{}} terminates and computes in $d$
the greatest common fixed-point of the functions from $F$.
\begin{proof}
  The proof is a direct adaptation of Apt's~\cite{AptTOPLAS2001}.  To
  prove termination it suffices to prove that the pair
  $(d,\#G)$ strictly decreases in some sense at each
  iteration of the \emph{\texttt{while}} loop, and to note that the
  ordering $\subseteq$ is well-founded.
  
  The correctness is implied by the invariant of the
  \emph{\texttt{while}} loop, i.e., every $f\in F-G$ is such that
  $f(d)=d$. It follows that the final domain is a common fixed-point
  of the functions from $F$ (since $G=\emptyset$).  The second part of
  the proof of Lemma~\ref{lem:fixed} ensures that it is the greatest
  one included in the initial domain.
\end{proof}
\end{theorem}

The following corollary concerns the application of a closure operator
in algorithm \GI{}.
\begin{corollary}\label{corol:idem}
  Consider operator $\phi$ that is applied in algorithm
  \emph{\GI{}}. If $\phi$ is idempotent, then assumption C
  is reduced to $\texttt{\emph{upC}}\,\texttt{\emph{:=}}\ \emptyset$.
\begin{proof}
See Apt~\cite{AptTOPLAS2001}.
\end{proof}
\end{corollary}

\subsection{On Properties of Functions}
In this section we examine the properties of reduction functions that
are preserved by compositions. We first define a set of properties
of interest for our purpose.
\begin{definition}
Consider two reduction functions $f,g$ on $\DOM$. Then, for all $x\in\DOM$:
\begin{itemize}
\item $f$ is idempotent if $ff(x)=f(x)$

\item $f$ and $g$ commute if $fg(x)=gf(x)$

\item $f$ semi-commutes with $g$ if $gf(x)\subseteq fg(x)$

\item $f$ is stronger than $g$ if $f(x)\subseteq g(x)$

\item $f$ and $g$ are independent if $fg(x)=gf(x)=f(x)\cap g(x)$

\item $f$ and $g$ are redundant if $f(x) = g(x)$

\item $f$ and $g$ are weakly redundant if $f\uparrow\omega(x) = g\uparrow\omega(x)$
\end{itemize}
\end{definition}

The property of idempotence of a composition operator allows us to
modify assumption C on the \texttt{update} function of the \GI{}
algorithm (see Corollary~\ref{corol:idem}). For this purpose 
some kinds of composition operators are shown to be idempotent
in the following proposition.
\begin{proposition}
Consider a set of composition operators $\Phi=\{\phi_1,\dots,\phi_k\}$.
Then, $\phi$ is idempotent if:
\begin{enumerate}
\item[i)] $\phi$ is a closure operator $\Gfp{\Phi}$.

\item[ii)] $\phi$ is a sequence operator $\Seq{\Phi}$,
   $\phi_i$ semi-commutes with $\phi_{j}$ for $i>j$,
   and each $\phi_i$ is idempotent.

\item[iii)] $\phi$ is a sequence operator $\Seq{\Phi}$,
   for each $i$ there exists $j<i$ such that
   $\phi_j$ is stronger than $\phi_i$,
   and $\phi_1$ is idempotent.

\item[iv)] $\phi$ is a decoupling operator $\Par{\Phi}$,
   each $\phi_i$ is idempotent, and the $\phi_i$ are pairwise independent.

\end{enumerate}
\begin{proof}
It suffices to prove that in all these cases, $\phi$ is idempotent.
The proof is then completed by Lemma~\ref{lem:fp}, i.e., each
element computed by $\phi$ is a common fixed-point of the functions
from its generator.
\begin{enumerate}
\item[i)] The proof is obvious.

\item[ii)] See Apt~\cite{AptTOPLAS2001}.

\item[iii)] Since the relation of strongness is transitive, then
   $\phi_1$ is stronger than $\phi_i$ for $i\in[2,\dots,k]$.
   Now it suffices to prove that $\phi_j\phi_1\dots\phi_k(x)=
   \phi_1\dots\phi_k(x)$. We have
   $$
    \phi_1(y)=\phi_1\phi_1(y)\subseteq \phi_j\phi_1(y)\subseteq\phi_1(y)
   $$
   since $\phi_1$ is idempotent, $\phi_1$ is stronger than $\phi_j$,
   and $\phi_j$ is contracting. This ends the proof if we set $y=\phi_2\dots\phi_k(x)$.

\item[iv)] Let us prove that function $\cap_{i=1}^{k}\phi_i$ is idempotent.
   We prove by induction on $j$ that $\cap_{i=1}^{j}\phi_i$ is
   idempotent for $j=1,\dots,k$ and that it is independent on
   $\phi_{j+1}$ for $j=1,\dots,k-1$.  It holds for $j=1$ since by hypothesis, $\phi_1$ is
   idempotent, and $\phi_1$ and $\phi_2$ are independent.  Now fix
   $1<j<k$, and consider that $\varphi:\cap_{i=1}^{j}\phi_i$ is
   idempotent, and that $\varphi$ and $\phi_{j+1}$ are independent. We
   prove that function $\psi:\varphi\cap\phi_{j+1}$ is idempotent and
   independent on $\phi_{j+2}$. Given an element $x\in\DOM$, we
   have:
$$
\begin{array}{lcll}
  \psi(\psi(x))
  & = & \varphi(\varphi(x)\cap\phi_{j+1}(x))\cap\phi_{j+1}(\varphi(x)\cap\phi_{j+1}(x))\\
  & = & \varphi\varphi(\phi_{j+1}(x)) \cap \phi_{j+1}\phi_{j+1}(\varphi(x))  & \hbox{ independence }\varphi,\phi_{j+1} \\

  & = & \varphi\phi_{j+1}(x) \cap \phi_{j+1}\varphi(x)
  & \hbox{ idempotence }\varphi,\phi_{j+1} \\

  & = & (\varphi(x)\cap\phi_{j+1}(x)) \cap (\phi_{j+1}(x)\cap\varphi(x))
  & \hbox{ independence }\varphi,\phi_{j+1} \\

  & = & \psi(x)
  & \hbox{ commutativity of }\cap \\
\end{array}
$$
Then $\psi$ is idempotent. Now we prove that $\psi$ and $\phi_{j+2}$
are independent, i.e.,
$$
  (\varphi\cap\phi_{j+1})(\phi_{j+2}(x))
  = \phi_{j+2}((\varphi\cap\phi_{j+1})(x))
  = (\varphi\cap\phi_{j+1})(x)\cap\phi_{j+2}(x)
$$
It suffices to prove that each term of this formula is
equivalent to $\cap\phi_{i=1}^{j+2}(x)$. It obviously holds for
the last term. For the first term, we have:
$$
\begin{array}{lcll}
  (\varphi\cap\phi_{j+1})(\phi_{j+2}(x))
& = & \varphi\phi_{j+2}(x)\cap\phi_{j+1}\phi_{j+2}(x)\\
& = & \phi_{1}\phi_{j+2}(x)\cap\cdots\cap\phi_{j}\phi_{j+2}(x)\cap\phi_{j+1}\phi_{j+2}(x)
\end{array}
$$
The independence of each pair $(\phi_i,\phi_j)$ ends the proof. For the second term we have:
$$
\begin{array}{lcll}
  \phi_{j+2}((\varphi\cap\phi_{j+1})(x))
& = & \phi_{j+2}\varphi(x)\cap\phi_{j+2}\phi_{j+1}(x)\\
\end{array}
$$
It suffices to remark that $\phi_{j+2}$ and $\phi_{j+1}$ are
independent by hypothesis, and then to prove that $\phi_{j+2}$ and
$\varphi$ commute, i.e.,
$\phi_{j+2}\varphi(x)=\varphi\phi_{j+2}(x)$.
This result is easily proved by induction since
$$\phi_{j+2}\varphi(x)=\phi_{j+2}\phi_1\dots\phi_j(x)$$
by independence of each pair $(\phi_i,\phi_j)$, and then
$$\phi_{j+2}\phi_1\dots\phi_j(x)=\phi_1\phi_{j+2}\phi_2\dots\phi_j(x)$$
by independence of $\phi_{j+2}$ and $\phi_1$, and so on.
\end{enumerate}
\end{proof}
\end{proposition}

In the following proposition, we identify cases where the computation
of a closure of a set of composition operators can be improved,
according to properties of independence and redundancy of operators.
\begin{proposition}\label{prop:redun}
Consider a set of composition operator $\Phi$ and a composition
operator $\varphi$. Then, the following properties hold:
\begin{enumerate}
\item If for all $\phi\in\Phi$, $\varphi$ and $\phi$ are independent, then
  $\Gfp{(\Phi\cup\{\varphi\})} = \Gfp{\Phi} \cap \varphi\uparrow\omega$.

\item If there exists $\phi\in\Phi$ such that $\varphi$ and $\phi$ are
weakly redundant, then $\Gfp{(\Phi\cup\{\varphi\})} = \Gfp{\Phi}$.
\end{enumerate}
\begin{proof}
\begin{enumerate}
\item It suffices to show that operator $\theta:\Gfp{\Phi} \cap
  \varphi\uparrow\omega$ is idempotent, i.e., $\theta\theta(x)=\theta(x)$
  for all $x\in\DOM$. The proof then follows by Lemma~\ref{lem:fixed}.
  Given a particular element $x$, consider that
  $\theta(x)=\phi_{i_1}\dots\phi_{i_k}(x)\cap\varphi\dots\varphi(x)$,
  and that
  $\theta\theta(x)=\phi_{j_1}\dots\phi_{j_l}\theta(x)\cap\varphi\dots\varphi\theta(x)$.
  A simple induction, using the hypothesis of independence, then
  allows us to rewrite $\theta\theta(x)$ as $\theta(x)\cap\theta(x)$.
  
\item The proof is obvious since by definition, $\Gfp{\Phi}$ is a
  fixed-point of $\phi$.  Hence, it is also a fixed-point of
  $\varphi$.
\end{enumerate}
\end{proof}
\end{proposition}

\subsection{From Decomposition to Composition}
A glass-box solver mainly combines elementary solving components. The
relation between solvers and components is expressed by
Definition~\ref{def:decomp}.
\begin{definition}[Decomposition of a function]\label{def:decomp}
  Consider a reduction function $f$ on $\DOM$. A finite set of
  reduction functions $F$ on $\DOM$ is a \emph{decomposition} of $f$
  if $\#F>1$ and for every $x\in\DOM$, we have $\Gfp{F}(x) =
  \LIMF{f}(x)$.
\end{definition}
For instance consider a filtering algorithm enforcing a local
consistency technique over a CSP. The generic computation is an
iteration such that the consistency of a set of constraints, a
constraint, or a constraint projection is verified at each step. Note
that there is a correspondence between the decomposition of a data
(e.g., a CSP) in a set of elementary components (e.g., constraints),
and the decomposition of a function (e.g., a solver associated with a
CSP) in a conjunction of elementary functions (e.g., an elementary
solver associated with constraints).

Considering different levels of granularity in a constraint solving
process must not influence the semantics of computation.
To this end we have the following result.
\begin{proposition}\label{prop:decomp}
  Consider a finite set of reduction functions $F=\{f_0,f_1,\dots,f_k\}$
  on $\DOM$.  Let $G$ be a decomposition of $f_0$. Given
  $H=G\cup\{f_1,\dots,f_k\}$, we have $\Gfp{H}(x)=\Gfp{F}(x)$ for
  every $x\in\DOM$.
\begin{proof}
The proof is very similar to the one of Lemma~\ref{lem:stab}.
\end{proof}
\end{proposition}

The decomposition of reduction functions leads to the notion of
decomposition relation. It can be useful to compare levels of
granularity of iteration algorithms.
\begin{definition}[Decomposition relation]\label{def:decomprel}
  Consider $F,G$ two finite sets of reduction functions on $\DOM$.
  $G$ is a \emph{decomposition of $F$} if
  $\#G>\#F$ and $\Gfp{G}(d)=\Gfp{F}(d)$ for
  every $d\in\DOM$.
\end{definition}
Note that the decomposition relation is a strict partial order.

In practice the aim is to fastly design efficient algorithms. We
believe that these requirements can be achieved in a glass-box and
generic (e.g., object oriented) programming approach.
\begin{itemize}
\item Developing powerful propagation techniques can be tackled by the
  \GI{} algorithm taking as input a well-chosen decomposition of
  solvers.  The generation of composition operators during the
  iteration allows us to efficiently combine elementary solvers using
  the best strategy with respect to the knowledge or properties of
  reduction functions. Several efficient strategies, dynamic in
  essence, are described in Section~\ref{sec:spec}.
  
\item Efficient prototyping is achieved by a generic programming
  approach. In fact in the case a generic (constraint propagation)
  algorithm is re-used, only a (generic) piece of code for each kind
  of elementary solver has to be implemented, e.g., for a reduction
  function enforcing a local consistency property.  In practice a main
  challenge is then to automatically generate the set of reduction
  functions of the decomposition given a generic implementation of a
  function and a CSP.
This issue is discussed in the conclusion but a precise
description is out of the scope of this paper.
\end{itemize}

\section{Specialized Algorithms}\label{sec:spec}
The constraint propagation framework described in this paper
allows us to tackle existing, efficient algorithms based on
strategies and heuristics.

\subsection{Scheduling of Constraints with Priorities}
\texttt{Choco} is a constraint programming system for finite domains
that has been developed by Laburthe et al.~\cite{Choco}.  The core
algorithm is constraint propagation whose main feature is to process
constraints according to the complexity in time of their associated
solving algorithms. Thus, primitive constraints are first processed, and
then, global constraints with linear, quadratic complexity, and so on
until a fixed-point is reached. In other words the computation is a
sequence of application of closures, each closure being connected
to the previous one by means of propagation events (modification of
domains).

Implementing the propagation engine of \texttt{Choco} using our
generic algorithm \GI{} can be done by considering priorities of
reduction functions. The creation of composition operators can be
implemented in two ways as follows:
$$
\begin{array}{llll}
1.~~ & \phi \texttt{:=}\, \Gfp{\{g\in G\mid \hbox{\emph{priority}}(g) = \alpha\}} & \hbox{s.t. } &
    \alpha = \min(\{priority(g)\mid g\in G\})\\
2. & \phi \texttt{:=}\, \Seq{(\Gfp{G_1}\cup\cdots\cup\Gfp{G_p})} & \hbox{s.t. } &
  G_1,\dots,G_p \hbox{ is a partition of } G\\
   &&& \forall i\in [1,\dots,p], \forall g\in G_i, \mathrm{priority}(g) = \alpha_i\\
  &&& \alpha_p < \alpha_{p-1} < \cdots < \alpha_1
\end{array}
$$
The first implementation describes the computation of the closure
of the set of active reduction functions with the greatest priority
(i.e., the computationally less expensive functions). It is a model of
\texttt{Choco} in the sense that only functions with greatest priority
are applied at a time. Note that the composition operator is a closure
(and thus, idempotent), and consequently Corollary~\ref{corol:idem}
applies for the \texttt{update} function (i.e.,
\texttt{upC}$=\emptyset$).

The second implementation is a sequence of closures, each closure
processing the set of functions of a given priority. The main
difference with respect to the first method is that no propagation
step is performed between the application of two closures. This
approach tends to minimize costs for updating propagation structures.

\subsection{Sequentiality in Interval Constraints}
The solver presented in~\cite{GranvilliersRC2001} implements
constraint propagation for interval domains, where reduction functions
enforce box consistency for numeric constraints over the reals. It
extends two existing solvers, namely \texttt{Numerica}~\cite{Numerica}
by Van Hentenryck et al., and the algorithm
\texttt{BC4}~\cite{BenhamouICLP1999} by Benhamou et al.

The solving process combines three kinds of reduction functions:
\begin{enumerate}
\item Function $f_{c,i}$ computes box consistency for the domain of variable
  $x_i$ with respect to constraint $c$, i.e., $f_{c,i}$ is a projection function;

\item Function $g_c$ implements constraint inversion for all variables
  of constraint $c$;

\item Function $h$ computes a linear relaxation (by means of a first order
Taylor approximation) for a constraint system, which is then processed
by the interval Gauss-Seidel method.
\end{enumerate}
The best strategy is based upon properties and heuristics: function
$f_{c,i}$ is stronger than the projection of $g_c$ on $x_i$, but they
are redundant if $x_i$ occurs only once in $c$; $g_c$ is
computationaly less expensive than functions $f_{c,i}$ for all $i$;
$h$ is in general more precise for tight domains, while the $f_{c,i}$
and $g_c$ are more efficient for large domains.
Using \GI{} this strategy is efficiently implemented as follows:
$$
\phi \texttt{:=}\, \Seq{(\Gfp{(G-\{h\})}\cup\Gfp{\{h\}})}
$$
Note that $h$ depends on all variables, that implies that $h$
belongs to $G$.

Furthermore, the decomposition process that generates the set of
reduction functions is tuned with respect to the redundancy property.
Hence, each reduction function $f_{c,i}$ such that $x_i$ occurs only
once in $c$ is removed (since $g_c$ is as precise as $f_{c,i}$ for $x_i$).
Proposition~\ref{prop:redun} guarantees that the output domain is
consistent with respect to all constraints from the initial system.

\subsection{Acceleration of Interval Narrowing}
Interval narrowing, i.e., constraint propagation with interval
domains, is inefficient if slow convergence happens. A slow
convergence corresponds to a cycle of reduction functions $f_i\dots
f_j\dots f_i$ such that each application only deletes a small part of
a domain. When constraints are nonlinear constraints over the
reals, this problem frequently arises, due to, e.g., singularities or points
of tangence.

Lhomme et al.~\cite{LhommeJLP1998} have devised an efficient strategy
based on cycle detection and simplification. The aim is to locally
select and apply the best reduction functions while delaying some
active functions supposed to slow the computation. Given a cycle,
i.e., a set of functions $G'$ from the propagation set $G$, the
solving process using \GI{} can be described as follows:
\begin{enumerate}
\item $G'$ is rewritten as $\Phi\cup\Phi'$ where $\Phi'$
  contains all functions $g$ from $G$ such that for all
  $\phi\in G-\{g\}$, $g$ and $\phi$ are independent;
  
\item $\Phi$ is rewritten as $\Phi_1\cup\Phi_2$, $\Phi_1$ being
  composed of the best functions from $\Phi$. More precisely $\Phi_1$
  contains a function $f$ per variable whose (current) domain can be
  modified by a function from $\Phi$; the selected function $f$ is the
  one that computes the largest reduction;

\item Doing so, the composition operator applied in \GI{}
  can be defined as:
  $$
  \phi \texttt{:=}\, \Seq{(\Gfp{\Phi_1}\cup\Phi_2)}
  $$
  applying first the best functions, and then the ones that have
  been delayed because of the independency property.
  Note that each function from the set $(G-G')\cup\Phi_2$ has to be
  added in $G$ by the \texttt{update} function.
\end{enumerate}

\subsection{Parallel Constraint Propagation}
Parallel processing of numerical problems via interval constraints has
been proposed as a general framework for high-performance numerical
computation in~\cite{HainsPacrim1997}. Parallel constraint
propagation~\cite{MonfroySAC1999} operationally consists in
distributing reduction functions among processors, performing local
computations, and then accumulating and intersecting new domains on
some processors.

The decoupling composition operator can be used to implement parallel
constraint propagation. A basic strategy is to create a partition
of the propagation structure $G = G_1\cup\dots\cup G_k$, $k$ being
dependent on the number of processors, and to consider operator $\phi$
to be applied in the algorithm \GI{}:
$$
\phi \texttt{:=}\, \Seq{G_1}\cap\cdots\cap\Seq{G_k}
$$
Moreover if one wants to perform more local computations before
synchronisation and communication, a closure can be computed
on each processor as follows:
$$
\phi \texttt{:=}\, \Gfp{G_1}\cap\cdots\cap\Gfp{G_k}
$$

Nevertheless, it has been observed that the classical notion of
parallel speed-up is not a correct measure of success for such
algorithms. This is due to a parallel decoupling phenomenon:
convergence may be faster when two interval contractions are applied
in sequence than in parallel. As a consequence a parallel version of
Lhomme's strategy, described in the previous section, has been
proposed in~\cite{GranvilliersIPL2000}. Essentially, parallelism is
only used to select the best functions, i.e., to create
the decomposition $\Phi_1\cup\Phi_2$ of the previous section.

Note that the decoupling phenomenon can be controlled according to the
independence property of reduction functions.
Proposition~\ref{prop:redun} ensures that a closure can be computed in
parallel, provided that each subset of dependent functions is located
on one processor. This property is not achievable in general for the
whole CSP. Nevertheless the number of links between processors
(corresponding to couples of dependent functions) can be minimized,
which tends to maximize the amount of reductions of domains. In that
case it may also be efficient to duplicate some functions on several
processors in order to break some links. Further work will
investigate such strategies, which have to be dynamic to guarantee load
balancing.




\section{Conclusion and Perspectives}\label{sec:conclu}
A set of composition operations of reduction functions is introduced
to design dynamic constraint propagation strategies. K.~R.~Apt's iteration
model is slightly modified while preserving the semantics.  Finally,
several well-known strategies (using priorities of constraints,
heuristics on the order of application of functions, and parallelism)
are modeled using a single iteration algorithm.

A generic implementation of constraint propagation, integrating
composition operators, has been designed. However it is out of scope
of this article and it will be the topic of a second article.

The set of composition operators is (intentionally) reduced to
sequence, closure, and decoupling operators. One may desire additional
operators to model sequences of fixed length, quasi closures with a
notion of precision, or conditional strategies with respect to dynamic
criteria. We believe that their integration in our framework
is feasible.

\subsection*{Acknowledgements}
We are grateful to Fr\'ed\'eric Benhamou for interesting discussions on
these topics.

\bibliographystyle{plain}

\end{document}